\documentclass[twocolumn,aps,prb, unsortedaddress, superscriptaddress]{revtex4-1} 

\usepackage{fancyhdr}		

\usepackage[section]{placeins}   %

\usepackage{graphicx}

\makeatletter \renewcommand\@biblabel[1]{$^{#1}$} \makeatother
\usepackage[mathlines]{lineno}

\usepackage{hyperref}
\hypersetup{ colorlinks,
    citecolor=blue,
    filecolor=blue,
    linkcolor=blue,
    urlcolor=blue
}

\usepackage{xcolor}

\definecolor{gray}{rgb}{0.6,0.6,0.6}
\definecolor{red}{rgb}{0.85,0,0}
\definecolor{green}{rgb}{0,0.85,0}
\definecolor{blue}{rgb}{0,0,0.85}
\definecolor{beige}{rgb}{0.92,0.87,0.78}
\usepackage[all]{hypcap}    

\begin{document}
\title{PYRO-NN: Python Reconstruction Operators\\ in Neural Networks}
\author{Christopher Syben}
\email[Email: ]{christopher.syben@fau.de}
\affiliation{Pattern Recognition Lab, Friedich-Alexander Universit\"at Erlangen-N\"urnberg, 91058 Erlangen, Germany}
\author{Markus Michen}
\affiliation{Pattern Recognition Lab, Friedich-Alexander Universit\"at Erlangen-N\"urnberg, 91058 Erlangen, Germany}
\author{Bernhard Stimpel}
\affiliation{Pattern Recognition Lab, Friedich-Alexander Universit\"at Erlangen-N\"urnberg, 91058 Erlangen, Germany}
\author{Stephan Seitz}
\affiliation{Pattern Recognition Lab, Friedich-Alexander Universit\"at Erlangen-N\"urnberg, 91058 Erlangen, Germany}
\author{Stefan Ploner}
\affiliation{Pattern Recognition Lab, Friedich-Alexander Universit\"at Erlangen-N\"urnberg, 91058 Erlangen, Germany}
\author{Andreas K. Maier}
\affiliation{Pattern Recognition Lab, Friedich-Alexander Universit\"at Erlangen-N\"urnberg, 91058 Erlangen, Germany}

\pagenumbering{roman}
\setcounter{page}{1}
\pagestyle{plain}

\begin{abstract}
\noindent {\bf Purpose:} Recently, several attempts were conducted to transfer deep learning to medical image reconstruction. An increasingly number of publications follow the concept of embedding the CT reconstruction as a known operator into a neural network. However, most of the approaches presented lack an efficient CT reconstruction framework fully integrated into deep learning environments. As a result, many approaches are forced to use workarounds for mathematically unambiguously solvable problems.\\
{\bf Methods:} PYRO-NN is a generalized framework to embed known operators into the prevalent deep learning framework Tensorflow. The current status includes state-of-the-art parallel-, fan-  and cone-beam projectors and back-projectors accelerated with CUDA provided as Tensorflow layers. On top, the framework provides a high level Python API to conduct FBP and iterative reconstruction experiments with data from real CT systems.\\
{\bf Results:} The framework provides all necessary algorithms and tools to design end-to-end neural network pipelines with integrated CT reconstruction algorithms. The high level Python API allows a simple use of the layers as known from Tensorflow. To demonstrate the capabilities of the layers, the framework comes with three baseline experiments showing a cone-beam short scan FDK reconstruction, a CT reconstruction filter learning setup, and a TV regularized iterative reconstruction. All algorithms and tools are referenced to a scientific publication and are compared to existing non deep learning reconstruction frameworks. The framework is available as open-source software at \url{https://github.com/csyben/PYRO-NN}.\\
{\bf Conclusions:} PYRO-NN comes with the prevalent deep learning framework Tensorflow and allows to setup end-to-end trainable neural networks in the medical image reconstruction context. We believe that the framework will be a step towards reproducible research
and give the medical physics community a toolkit to elevate medical image reconstruction with new deep learning techniques. \\

\end{abstract}
\maketitle



\pagenumbering{arabic}
\setcounter{page}{1}
\pagestyle{fancy}
\section{Introduction}

In recent years, major breakthroughs made deep learning increasingly prevalent in
more and more fields. It revolutionizes the way of classification and regression
tasks in speech and image recognition \cite{lecun2015deep, imagenet, wavenet} and
many other areas. Even in the medical domain, where interpretability and
reliability are one of the most important driving forces, deep
learning has led to astonishing results \cite{maier2018gentle}. One
of the most cited papers of recent years is the U-net \cite{unet} which
outperforms classical machine learning algorithms in segmentation tasks. In the
subsequent time, the U-net architecture emerged to many more tasks, e.g., artifact
correction, image fusion, image-to-image translation, and even into the context of
medical image reconstruction \cite{stimpel2017mr, jin2017deep, kofler2018u}. However, this domain is fundamentally different from
those in which the advent of deep learning began, and the question arises as to
whether these learned signal reconstruction pipelines are reliable and stable enough for a
critical area such as medical imaging \cite{instabilities}.
Two special issues: '\textit{Deep learning in medical imaging}' \cite{tmi1} and
'\textit{Machine Learning for Image Reconstruction}' \cite{tmi2} in Transactions
on Medical Imaging (TMI) in 2016 and 2018 discuss the increasing relevance of deep
learning methods in medical image reconstruction. 

The presented approaches can be divided into either pre- or post-processing approaches or fully end-to-end trained methods. For the first type the actual reconstruction pipeline is based on well known signal reconstruction algorithms omitting the end-to-end capability due to its complexity. For the second type of approaches the modeling of the end-to-end pipeline can be realized under two different paradigms.
One way is to learn the whole signal processing pipeline, an exceptionally clear representative of this paradigm is \mbox{\textit{AUTOMAP}}~\cite{automap}. Directly in contrast to this is the emerging paradigm of embedding known operators \cite{precisionLearning}. This preserves the end-to-end learning capability but includes the known operations of the reconstruction chain to preserve the credibility of the signals, reduce the error bound of the learning process and decrease the number of parameters and thus the amount of necessary training data. This paradigm gets increasingly popular, with multiple publications following the way of embedding known operators in the Computed Tomography (CT) context and successfully including the CT reconstruction as known operators into the network architecture to be able to benefit from the end-to-end training capability of deep learning \cite{ye2018deep, chen2018learn,alderPrimalDual,wurflmiccai,hammernikbvm,sybenctmeeting,wurfltmi,sybengcpr}. 
However, the publications that follow this path are still less represented than those that use deep learning only as pre- or post-processing. 
We believe that a major reason for this is the non-trivial implementation of known operators in existing deep learning frameworks. Even publications that successfully take on this challenge often refer to their own implementations as prototypical\cite{chen2018learn} or provide frameworks on abstract wrapped levels\cite{alderPrimalDual,odl}. An efficient and publicly usable solution integrated into on of the popular deep learning frameworks, however, remains pending.

To strengthen the paradigm of known operators, elaborate the research in the medical image reconstruction, and to avoid reimplementations and incompatibilities, we started to work on an open source software framework PYRO-NN, which allows an easy way to integrate known algorithms into the deep learning framework Tensorflow \cite{tensorflow}. We provide multiple forward and backward projectors for CT implemented in CUDA based on scientific publications supported with a high level Python API for simple use of state-of-the-art CT reconstruction, even from different setups of real CT scanners. The profound integration into Tensorflow on C++/Cuda level allows to handle occurring performance and memory issues and, additionally, allows an easy customization of the algorithms compared to a wrapper alternative like \cite{Astra,odl}. Furthermore, the high level Python API offers an easy link between deep learning and community driven frameworks. For the CT domain this allows to use a wide range of tools (e.g. filter, redundancy weights, etc.)~\cite{Astra, tomopy, Conrad, pyconrad}.

We believe that this framework will help the community leverage the power of end-to-end training of machine learning algorithms directly from the data, while continuing to apply mathematically sound solutions to uniquely solvable problems.

\section{Methods}
The framework concept is designed to easily include C++ and CUDA based algorithms into the deep learning framework Tensorflow. In detail, PYRO-NN provides network layers as CUDA implementations to generate parallel-, fan-, and cone-beam X-ray projections and to reconstruct them within any neural network constructed with Tensorflow. Due to the nature of the projection and reconstruction operation we intrinsically provide the analytical gradients for all of these layers, which allows fully end-to-end trainable networks. Furthermore, with PYRO-NN we provide filters and weights based on scientific publications to allow proper filtered-backprojection (FBP) reconstructions. The PYRO-NN API is inspired by the CONRAD \cite{Conrad} framework to adapt the ability to reconstruct data from real clinical scanners and by using PyConrad \cite{pyconrad} many more tools and phantoms can easily be used in the deep learning context. The current state of the framework features a CT reconstruction pipeline, while the basic design allows to transfer the whole concept to other signal reconstruction domains within one framework and, therefore, points out a  direction to future development and community contribution.

This section is structured as follows: First we introduce the general design of the framework, followed by a description of how to create the link to the Tensorflow framework. In the second part we introduce the operator and algorithms necessary to conduct CT reconstruction and describe the two levels of the framework: the actual layer implementation on the C++/CUDA level and the high level Python API providing a convenient usage of the provided layers.

\subsection{Software Design / Rationale }
\begin{figure*}[tb]
	\centering
	\includegraphics[width=0.55\textwidth]{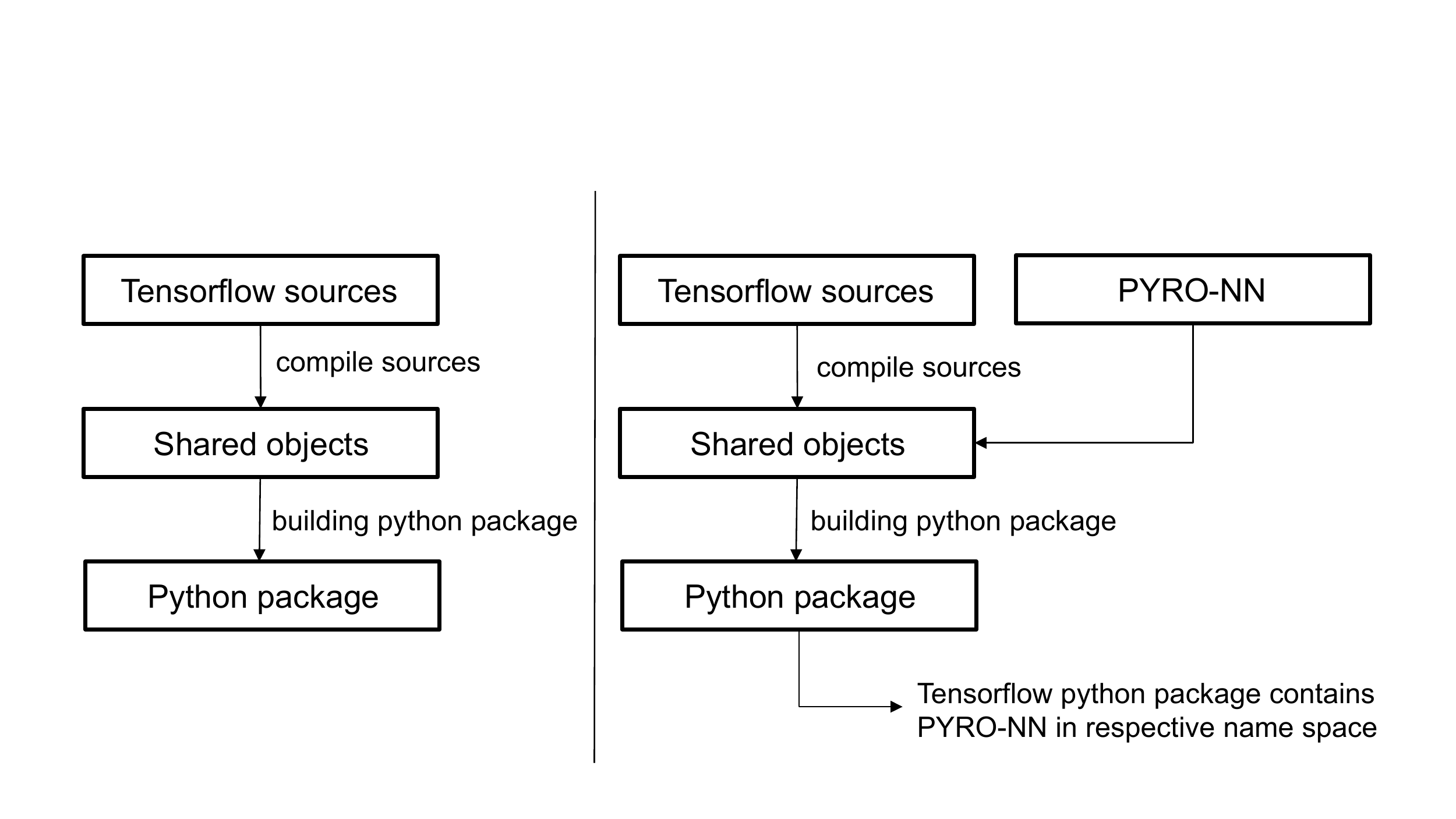}
	\caption{The build process. Left the build process defined by Tensorflow, right our adjusted process.}
	\label{fig:built_process}
	\vspace{0.2cm}
\end{figure*}

The development speed in the deep learning community is tremendous. Like in the research itself, the toolkits and frameworks are developing in the same speed, which often causes conflicts in interoperability of self-developed solutions and version mismatches between different frameworks and toolkits. To ensure a robust version control, the framework is directly included into the building process of the Tensorflow sources. In Fig. \ref{fig:built_process} the process is shown. We provide a patch for all Tensorflow versions since 1.9 which adjust the build process of the Tensorflow sources such that all C++ and CUDA files in the framework directory are built and included into the Tensorflow Python package under the PYRO-NN namespace. This yields the benefit that the user can define specific version combinations for their purpose. Furthermore, the separation of the custom implemented algorithms and the information flow allows the possibility to extend support of the proposed framework towards other deep learning frameworks like PyTorch.

\begin{figure*}[tb]
	\centering
	\includegraphics[width=0.65\textwidth]{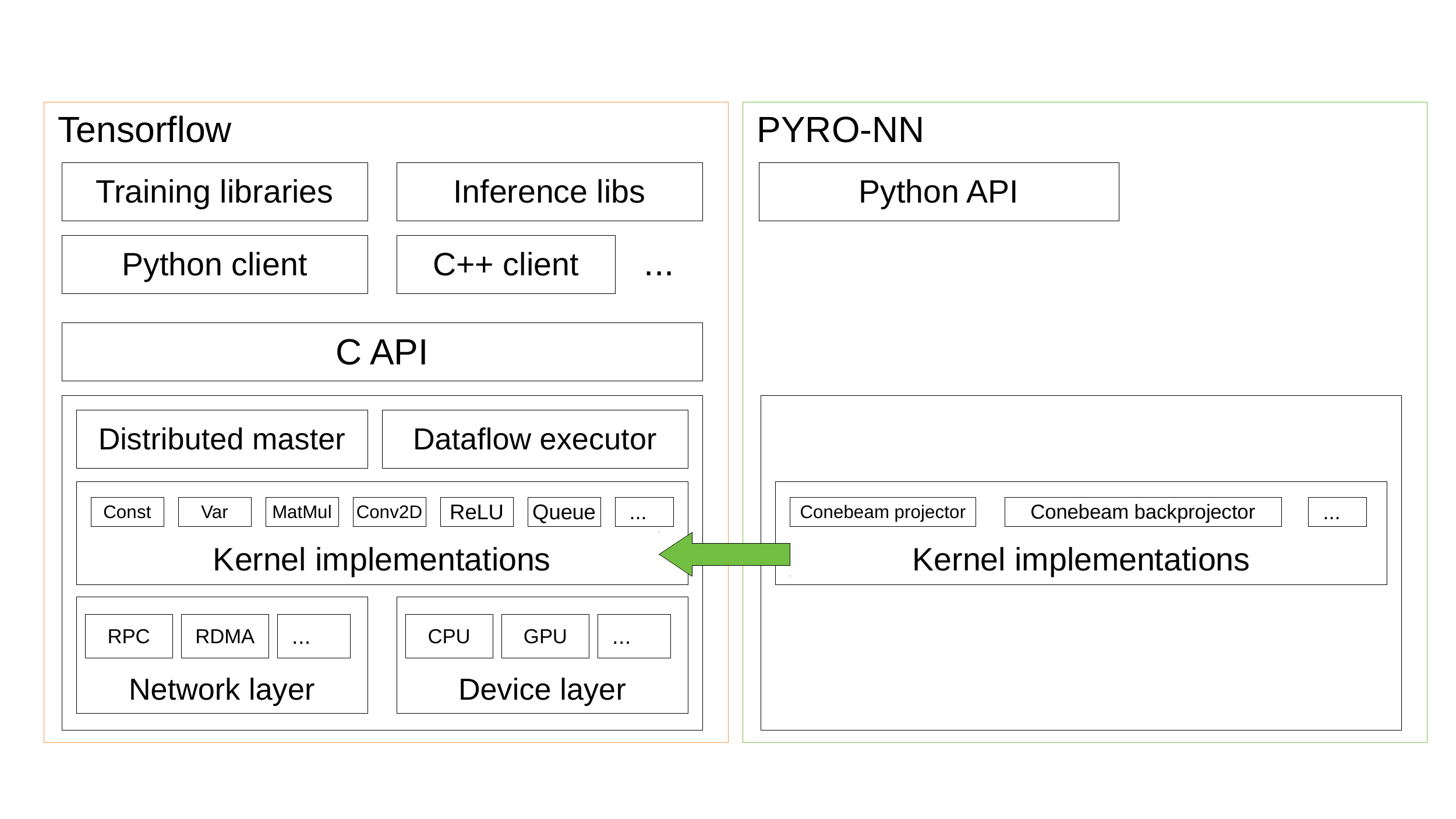}
	\caption{The architecture of PYRO-NN beside to Tensorflow. Image adjusted from https://www.tensorflow.org/guide/extend/architecture .}
	\label{fig:architecture}
	\vspace{0.2cm}
\end{figure*}

The architecture is mainly defined by the API to Tensorflow. In Fig. \ref{fig:architecture} the abstract concept is visualized. Known operators can be implemented as kernels consisting of a C++ class following the design of the Tensorflow API and the actual implementation of the operator as a kernel. The C++ class ensures the behavior of the kernel as a Tensorflow layer in the neural network. In contrast to other frameworks which wrap the implementation on the Python level, this gives full control over the device resources like memory usage.

The third part is the high level Python API. Due to the API defined by Tensorflow the user is able to use the known operators like normal Tensoflow layers with the defined input, output and attributes. A crucial aspect is the registration of the gradient to the respective layer. This allows to invoke the relevant algorithm to compute the gradient with respect to the layer in an efficient way. The provision of the gradient is a necessity to enable a gradient flow through the entire network and, thus, allow fully end-to-end trainable networks with known operators. The high level Python API allows to design helper structures for the respective operator. For the CT context, we provide wrapped layers with automated gradient registration and object oriented classes to feed the known operators the necessary attributes. The class design follows the high level abstraction of Tensorflow and wraps the implementation detail to provide the user an easy and convenient usage while preserving the possibility for the user to modify the wrapped methods.

All together, these rationals offer the community with a generic, version stable, framework to easily include known operators into neural networks. The source code is publicly available under the Apache 2.0 licence to be directly compatible with Tensorflow and to allow an uncomplicated integration of the community into existing projects. In the following the implementation of these rationals are shown on our case study of CT reconstruction, which provides a broad band of tools to allow end-to-end networks with reconstruction capability.

\subsection{CT Reconstruction in Neural Networks}
Based on the generic design of the framework, the current state provides all necessary algorithms and tools for analytical parallel-, fan- and cone-beam reconstruction. The necessary algorithms are implemented within Tensorflow as an own layer, while the respective tools, e.g., filter,  weights, etc. are provided on the Python level to supply a high level API for CT reconstruction.
In the following, we introduce the mapping of the known operator to a layer, followed by a description of the provided algorithms and tools.

\subsubsection{The Known Operator}
For the task of reconstructing object information from acquired X-ray projections, an efficient analytical method is well known and is called Filtered-Backprojection (FBP). To embed these methods into a neural network, the whole acquisition and reconstruction procedure of a CT system needs to be described with discrete linear algebra to embed them into a neural network. The acquisition of projection data of the object can be described with
\begin{equation}
\mathbf{A}\mathbf{x} = \mathbf{p} \enspace,
\end{equation}
where $\mathbf{A}$ is the matrix describing the geometry, the so called system matrix which can be algorithmically implemented as the forward-projection operator. $\mathbf{x}$ is the object and $\mathbf{p}$ are the acquired projections of object $\mathbf{x}$ under the system described by $\mathbf{A}$. The reconstruction can be obtained with
\begin{equation}
\mathbf{x} = \mathbf{A^{-1}}\mathbf{p} \enspace,
\end{equation}
where $\mathbf{A^{-1}}$ is the inverse system matrix, which cannot be inverted because it is a tall matrix. Thus, the reconstruction can be conducted using the Moore-Penrose Pseudo Inverse
\begin{equation}
\mathbf{x} = \mathbf{A^\top}(\mathbf{AA^\top})^{-1}\mathbf{p}  \enspace,
\end{equation}
where $\mathbf{A^\top}$ is the adjoint system matrix which can algorithmically implemented as the back-projection operator. Following the analytical filtered back-projection approach, the inverse bracket with $\mathbf{A}\mathbf{A^\top}$  must be a filter operation with the ramp filter. The filtering step can be described by a multiplication with a diagonal matrix $\mathbf{K}$ in the Fourier domain
\begin{eqnarray}
\centering
\mathbf x = \mathbf A^\top \mathbf F^{\mathbf H} \mathbf K \mathbf F \mathbf p \enspace,
\label{eq:reco_matrix}
\end{eqnarray}
where $\mathbf{F}$, $\mathbf{F}^{\mathbf H}$ is the Fourier transform and the respective adjoint, i.e.~inverse operation. As Eq.~\ref{eq:reco_matrix} shows, the reconstruction of an object from projection data can be expressed completely as discrete linear algebra and, therefore, each matrix multiplication can be modeled as a layer in a neural network. The working horse of the neural network performance is based on the backpropagation algorithm to update the weights and, thus, train the network. To allow the integration of the presented operator, we need to ensure that the operator has at least a sub-gradient, to ensure the capability to propagate gradients through the whole network. The publications from W\"urfl and Syben et al. show that $\mathbf{A}$ and $\mathbf{A^\top}$ are their respective operators to calculate the gradient, thus the gradient flow through these layers can be ensured\cite{wurflmiccai,sybenctmeeting}. 

\subsubsection{The Operator as a Layer}

From iterative reconstruction it is well known that the system matrix does not fit into memory, therefore, we compute the operator on the fly using ray-based algorithms. There are several ways for the computation. We introduce the \textit{ray-driven forward-projection} and the \textit{voxel-driven back-projection} algorithmically with respect to the integration into Tensorflow. Note that when using a ray-driven forward-projection algorithm to compute the result of the multiplication with $\mathbf{A}$, then the voxel-driven back-projection algorithm is not the respective adjoint operation $\mathbf{A^\top}$. They are a so called a unmatched projector-/back-projector pair. The implications of matched projectors and shear-warp projectors on the convergence and runtime are subject to future work and are briefly discussed in Section \ref{discussion}

The \textit{forward-projection} to generate projections from the input volume are implemented as CUDA kernels in a ray driven manner. For each detector pixel, a ray $\vec{r}$ is cast through the scene, accumulating the absorption values along the line. We provide forward projectors for 2D parallel- and fan-beam geometry based on ray vectors and respective geometry parameters. Furthermore, a 3D cone-beam forward projector based on projection matrices is implemented according to Galigekere et al.~\cite{forwardprojection}. The projection matrices allow a simple generation of projection data under the geometry of real systems. The CUDA kernels are parallelized over the detector pixels computing the line integral along the ray. For the 3D cone-beam case the framework offers the possibility to choose between a texture and a kernel interpolation mode. While the texture interpolation of the graphics card leads to very short calculation times, the memory management of Tensorflow and the necessity of a pitched memory structure of CUDA for the data to be interpolated, which leads to the point that the volume to be projected is kept in the memory twice. To provide the community with the possibility of including the projection operator into big networks and with the option for precise memory management, kernel interpolation is provided. The ray-casting algorithm using the kernel-based interpolation is much slower but directly works with the allocated tensors from Tensorflow.

The \textit{back-projection} operators to reconstruct simulated or real projection data are implemented as CUDA kernels in a voxel-driven manner. For each pixel/voxel to be reconstructed, the projection of the point on all projection images is accumulated. The framework provides the respective 2D parallel- and fan-beam back-projection algorithms based on geometry parameters and ray vectors. Following the forward projection, the 3D cone-beam back-projection is based on projection matrices according to Scherl et al.~\cite{backprojection}. This allows to reconstruct data from simulations and real systems as shown in the CONRAD framework \cite{Conrad}. The back-projection kernel is parallelized over the voxels projecting the respective position on the different detector coordinates interpolating the measured line integral. Like the 3D cone-beam forward-projection implementation, the back-projection kernel is implemented with a texture and kernel interpolation mode to give full control over the memory management to the user.

\subsection{High Level Python API}
To supply the community with an easy-to-use version
of the described layers, we provide the necessary structure and additional tools like filters, weights, phantoms, etc. within the Python framework. In the following, the outline of the necessary structure to evaluate the layers is shown, followed by a short introduction of the provided tools with their related publications to support a state-of-the-art reconstruction pipeline.

\subsubsection{Reconstruction \& Geometry}
The high level Python API wraps the provided reconstruction layers in Tensorflow. Thus, the framework registers the respective adjoint operation for the gradient computation automatically. All attributes necessary for the provided forward- and backward-projection layers are covered with a base geometry class and corresponding specialized derived classes. The geometry is defined by \textit{volume shape, volume spacing, detector shape, detector spacing, number of projections, angular range} and for fan- and cone-beam geometries the \textit{source to iso-center distance} (SID) and \textit{source to detector distance } (SDD). Additionally, the geometry contains the ray vectors (2D) or the projection matrices (3D) to describe the respective trajectory.

\subsubsection{Phantoms}
PYRO-NN contains a set of simple geometric objects, e.g. circle, ellipsoid, sphere and rectangles to easily create a more complex numerical phantom. Furthermore, the framework provides an analytical description of the 2D Shepp-Logan phantom \cite{shepp_phantom} as well as a 3D extension based on the CONRAD implementation\cite{Conrad}.

\subsubsection{Trajectories}
The trajectory describes the geometric scanner setup over the whole scan. For the 2D parallel- and fan-beam cases the trajectory is described by the central ray vector for each projection. For the 3D cone-beam case the trajectory is described by a set of projection matrices, which allows to use calibrated projection matrices from real scanner systems. Within the high level Python API we provide basic methods to compute the respective rays or projection matrices based on a given geometry. This general concept gives the user the option to easily use individual trajectories.
The open-source concept of the whole framework allows to contribute to the diversity of provided trajectories.

\subsubsection{Filters}
To allow a basic reconstruction in the context of neural networks, PYRO-NN provides the Ramp and Ram-Lak filter implemented according to Kak and Slaney\cite{kak}. The filters can be directly assigned as weights to a multiplication layer and, thus, can be interpreted as a multiplication with a diagonal matrix in the Fourier domain as shown in Eq.~\ref{eq:reco_matrix}.

\subsubsection{Geometric Correction Weights}
In order to support fan- and cone-beam reconstructions, the framework contains a cosine-filter to weight the each pixel in the sinogram with its distance to the X-ray source and, thus, correct fan- and cone-beam artifacts. It is implemented according to Kak and Slaney\cite{kak}.
Furthermore, to support short and super short scans redundancy weights are provided e.g. Parker\cite{parker} and Riess\cite{riess} weights.

\subsection{Network Architectures}
Following the paradigm of precision learning, different network architectures can be setup or even derived as shown by Syben et al.~\cite{sybengcpr}. We provide various examples within the framework to assist users in using the framework. Therefore, we present a simple network able to reconstruct short-scan cone-beam CT according to the Feldkamp-Davis-Kress (FDK) algorithm \cite{fdk}. Additionally, an example of learning the correct discretization of the ramp filter is shown.


\section{Example Applications}
We show three baseline experiments to outline the capability of the provided framework. This provides the code base for an easy entry into deep learning-based signal reconstruction and to support comparability within emerging methods. 

\subsection{FDK Reconstruction Network}
\begin{figure*}[tb]
	\includegraphics[width=1.0\linewidth]{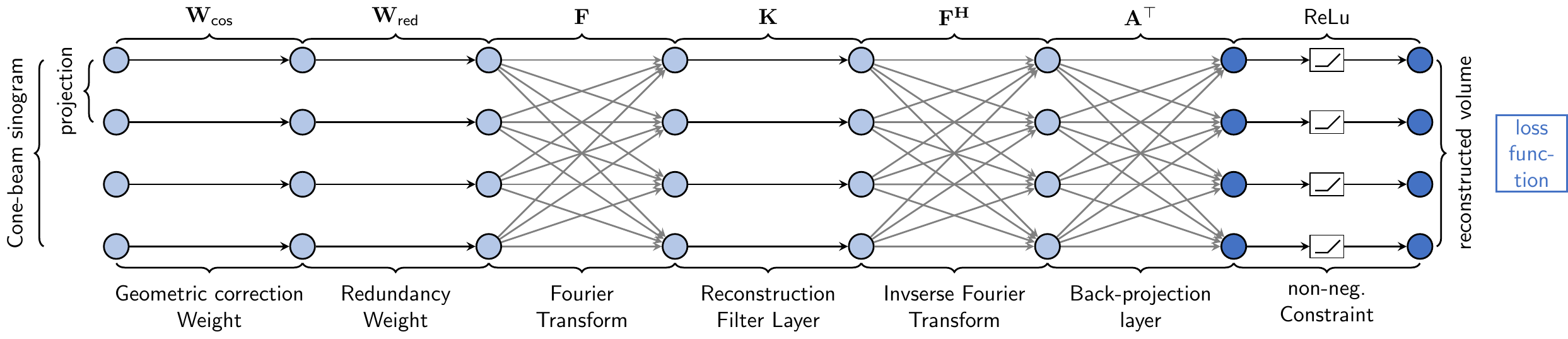}
	\caption{FDK-Reconstruction Network. Light blue nodes represent the projection
		domain, while dark blue nodes stands for volume domain.\newline\newline}%
	\label{fig:fdk_net}
\end{figure*}

\begin{figure*}[tb]
	\centering
	\includegraphics[width=0.7\linewidth]{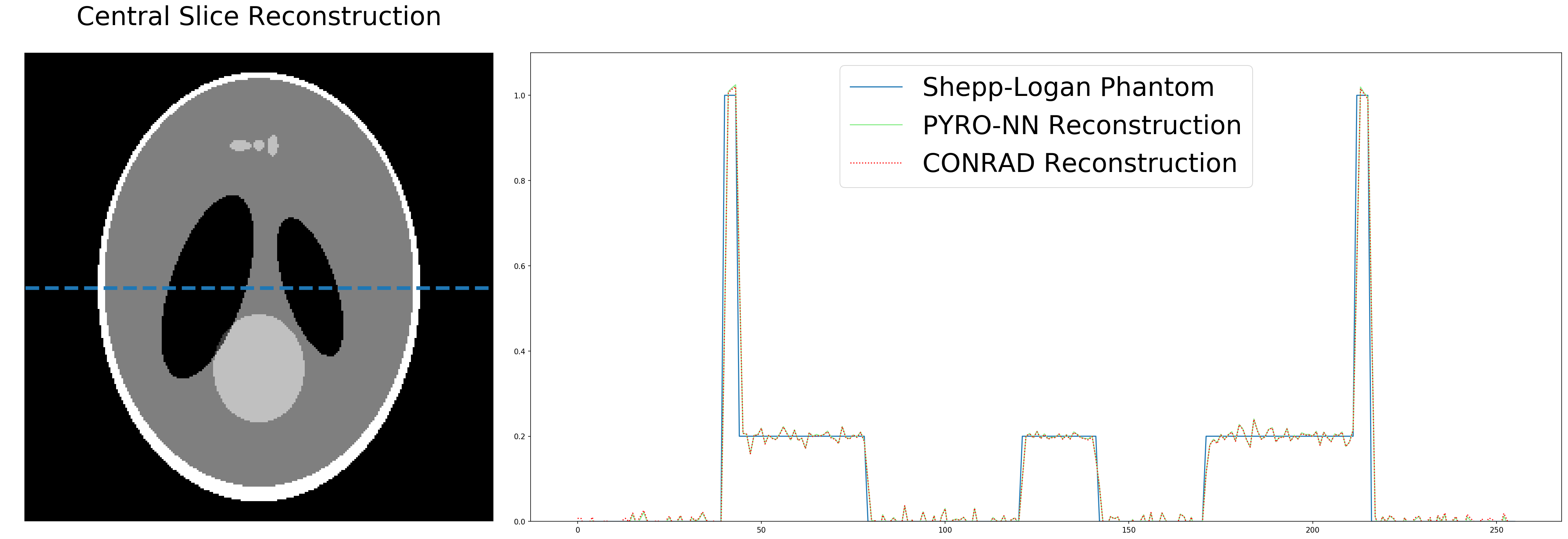}
	\caption{Short scan FDK reconstruction. The central slice reconstruction  shows the result of the network, comparing line plots from phantom (blue), PYRO-NN reconstruction (green) and reference CONRAD reconstruction (red). The gray-level window is [0,0.4].}%
	\label{fig:fdk_comparison}
\end{figure*}

As a proof of concept and validation of the method we provide a FDK reconstruction with Parker weights \cite{parker} for a short scan geometry formulated as neural network using the proposed Framework.
On the one hand, this setup serves to demonstrate the ability of the presented framework to perform a cone-beam reconstruction. On the other hand, the basic network is a good starting point which makes it possible to integrate further deep learning tools in projection or reconstruction domain. The whole pipeline can then be trained in an end-to-end fashion. 
This allows general data-driven optimization as well as task specific solutions. Note that the error e.g. $\ell2$-loss is computed in the reconstruction domain and can be used to update weights applied in the projection domain. The architecture is shown in Fig. \ref{fig:fdk_net}. The operation of the FBP algorithm for cone-beam short-scan geometry can be described with
\begin{equation}
\mathbf{x} = \mathbf{A^\top}\ \mathbf{F^{\mathbf{H}}}\ \mathbf{K}\ \mathbf{F}\ \mathbf{W}_{{red}}\ \mathbf{W}_{{cos}}\ \mathbf{p} \enspace,
\end{equation}
where $\mathbf{W}_{{cos}}$ is a diagonal matrix describing the pixel-wise independent weighting of the projection data $\mathbf{p}$ with cosine weights. For so called short-scan trajectories some rays are measured twice and some only once. These redundancies will lead to streaking artifacts in the reconstruction if the projection data is not adequately weighted. $\mathbf{W}_{{red}}$ weights the projection data appropriately, e.g., for a defined cone-beam short scan the weights by Parker \cite{parker} can be used.  $\mathbf{A^\top}, \mathbf{K}, \mathbf{F}$ and $\mathbf{F^{\mathbf{H}}}$ are the system matrix, reconstruction filter and Fourier transform as described in Eq.~\ref{eq:reco_matrix}.
The described network architecture can reconstruct a 200 degree cone-beam CT short-scan with 248 projections properly. In Fig.~\ref{fig:fdk_comparison} the reconstruction result of a 3D Shepp-Logan phantom is shown. The line profiles illustrate that the result follows the phantom and the reference FDK reconstruction using CONRAD \cite{Conrad}.

This architecture can be seen as a baseline, which can be used to insert deep-learning features and non-linearities in the projection as well as in the reconstruction domain while the gradient can be propagated through the whole network. For example, the redundancy weights can be learned for arbitrary short-scan angles. The update step for the weights according to the backpropagation algorithm is shown by W\"urfl et al. \cite{wurfltmi}. 

\subsection{Learn Reconstruction Filter}
To highlight the capability to propagate the gradient through the whole network, from reconstruction domain to projection domain, we demonstrate that the framework can be used to learn the discretized version of the analytical derived ramp-filter for reconstruction in an end-to-end network approach.
 
The network architecture is defined by
\begin{equation}
\mathbf{x} = \mathbf{A^\top}\ \mathbf{F^{\mathbf{H}}}\ \mathbf{K}\ \mathbf{F}\ \mathbf{p} \enspace,
\end{equation}
where $\mathbf{K}$ is the reconstruction filter initialized with the ramp filter and the only trainable layer in the network. For an filtered back-projection algorithm the ramp filter applied in the projection domain is the continuous analytical derived solution. Without proper discretization of the filter, offset and cupping artifacts will occur in the reconstructed volume. The proper discretization is the so called RamLak filter\cite{RamLak}.

\begin{figure*}[htb]
	\centering
	\includegraphics[width=0.6\linewidth]{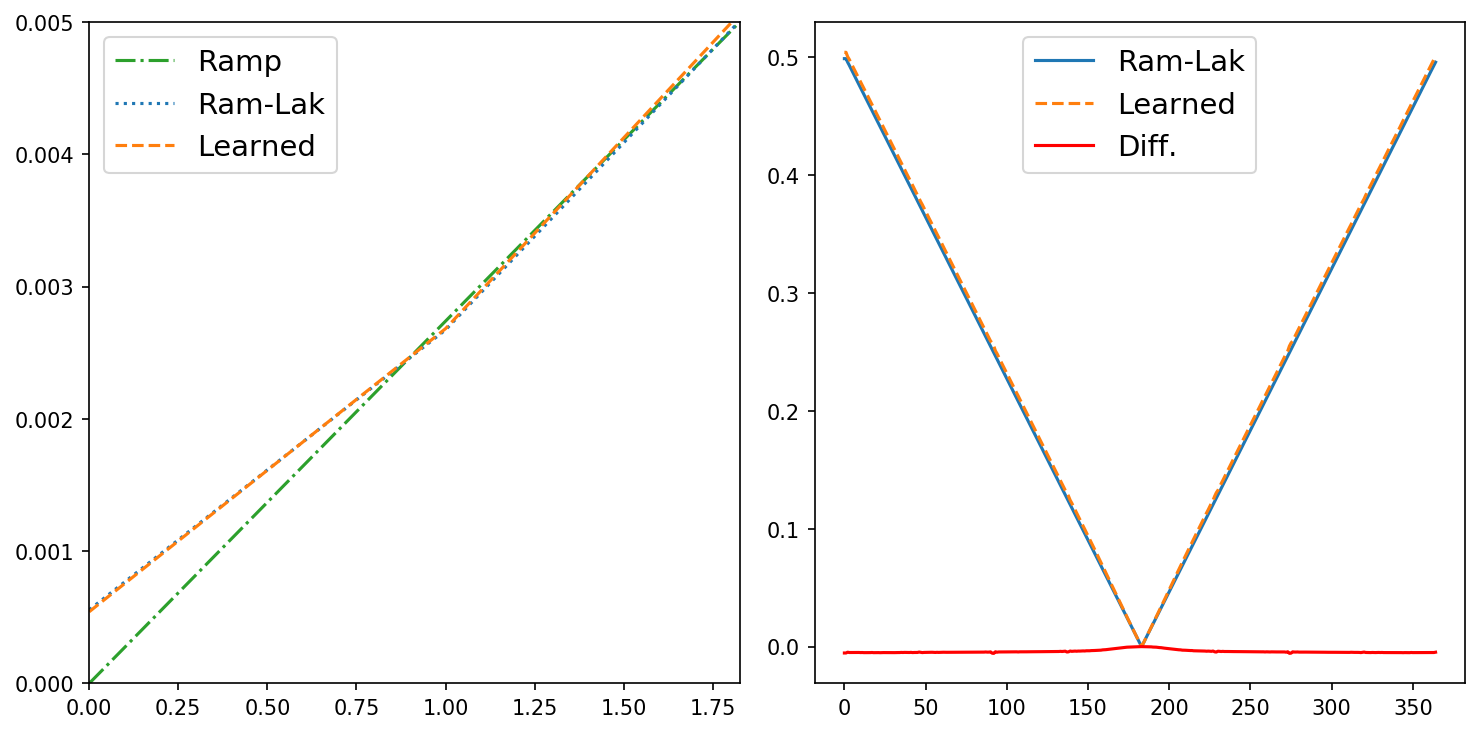}
	\caption{Ramp, RamLak and Learned filter.}%
	\label{fig:filer_learning}
\end{figure*}

We can show that we can use the backpropagation of the network to compute the error on the reconstruction domain and adjusting the filter weights of $\mathbf{K}$ in the projection domain. The result is converging towards the analytical derived discretized RamLak filter as shown in Fig.~\ref{fig:filer_learning}. A detailed description and discussion of the experiment is given by Syben et al.\cite{sybenctmeeting}. Note that the training of the filter was done with purely numerical data and can directly applied on real data without loss of validity.

\subsubsection{Iterative Reconstruction}
\begin{figure*}[htb]
	\centering
	\includegraphics[width=0.5\linewidth]{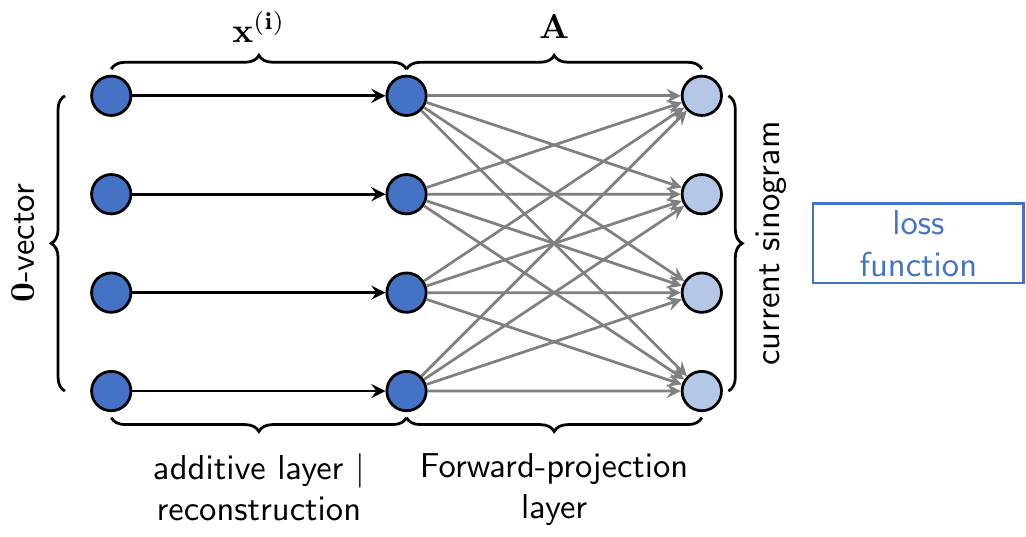}%
	\caption{Iterative reconstruction network. Light blue nodes represent the projection
		domain while dark blue nodes stand for volume domain.}%
	\label{fig:iter_reco}
\end{figure*}

With the third experiment we want to demonstrate the versatility and possibilities which are offered by the framework. We can model a network such that an iterative reconstruction algorithm with regularizers can be conducted. The network is designed such that the backpropagation algorithm mimics the iterative reconstruction based on a data fidelity term. Starting with the data fidelity term for iterative reconstitution with total variation (TV) regularizer
\begin{equation}
\min\ ||\mathbf{A}\mathbf{x} - \mathbf{p}||_2^2 + \lambda\mathbf{TV}(\mathbf{x})
\end{equation} 
where $\mathbf{A}$ is the system matrix, $\mathbf{x}$ is the volume to be reconstructed initialized with zero and $\mathbf{p}$ is the measured sinogram. The data fidelity term can be also interpreted as a network architecture, where the input of the network is a zero-vector, followed by an additive layer and then forward projected using the provided forward projection layer as shown in Fig.~\ref{fig:iter_reco}. The loss function is then the $\ell2$-loss w.r.t to the measured data $\mathbf{p}$. Training the network will update the weights of the additive layer. Thus, the seeked reconstruction is represented by the weights of the additive layer at the end of the learning process.  

\begin{figure*}[!htb]
	\centering
	\includegraphics[width=0.9\linewidth]{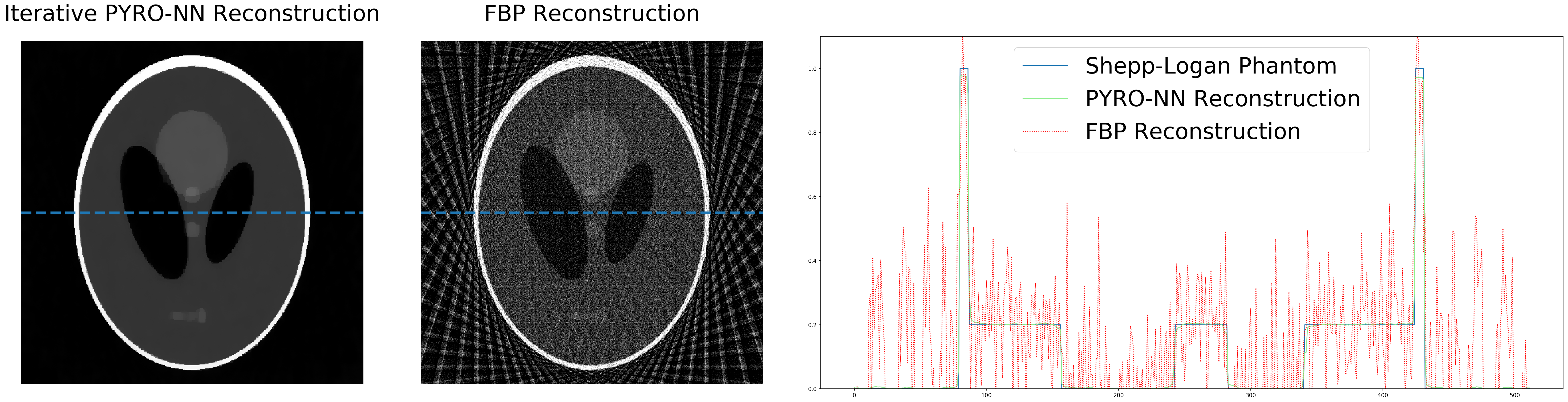}%
	\caption{Sparse view reconstruction. The reconstruction shows the result of the iterative reconstruction network, comparing line plots from phantom (blue), iterative PYRO-NN reconstruction (green) and reference FBP reconstruction (red). The gray-level window is [0,1.0].}%
	\label{fig:iter_comparison}
\end{figure*}

We demonstrate that the network setup can reconstruct a sparse view CT acquisition with 30 parallel projections over an angular range of 180 degree. 2\% additive Gaussian noise where applied to the projection data. In Fig.~\ref{fig:iter_comparison} the reconstruction with the iterative \mbox{PYRO-NN} reconstruction is shown and compared to an FBP reconstruction and the Shepp-Logan phantom. Note that the actual TV-regularized iterative reconstruction using the proposed framework needs only few lines of code. The computation of TV is provided by Tensorflow. With this baseline example we provide an easy-to-use starting point for future investigations into learned regularizers and different loss functions.

\section{Discussion}
\label{discussion}
Recently, there have been several different attempts to transfer the astonishing capability of deep learning into the field of medical image reconstruction. The approaches span from purely CNN based image-to-image translation task as pre- or post processing steps over end-to-end networks learning the complete signal reconstruction pipeline in a data-driven manner up to end-to-end networks which are enriched with known operators to preserve interpretability and reliability of the signals. The different downsides of these approaches were discussed in the community and range from adversarial attacks and signal falsification \cite{instabilities,miccaiAttack} over the need of huge amounts of data up to highly academic implementations not applicable on real scenarios due to huge memory requirements. In order to transfer deep learning towards medical image reconstruction and at the same time address these problems, the idea of embedding known operators into the neural networks is increasingly pursued as the growing number of publications shows. 

In this paper, we present a framework providing the known operators for CT reconstruction and all necessary tools to conduct experiments on real scenarios. We believe that such an open-source framework will reduce the barriers of such approaches and will elevate the research in the medical image reconstruction domain. To encourage the research we provide the shown baseline experiments as example code within the framework, allow an starting point for own research ideas. To the best of our knowledge there exist no framework that allows simple use of CT reconstruction algorithms within neural networks and can handle and reconstruct data from real systems. 

In order to show the capability of the provided framework, we presented a state-of-the-art FDK reconstruction using projection matrices from a real C-arm Cone-beam CT system and deliver this as one baseline example within the framework. This can directly be used to reproduce the results of W\"urfl et al.~\cite{wurfltmi}. Furthermore, we provide a reference implementation to learn CT reconstruction filters following Syben et al.~\cite{sybenctmeeting}. We want to emphasize that the learning process is based on purely numerical data without loss of generality. Both experiments show the capability to compute the error in the reconstruction domain while the weight updates can be applied for operators in the projection domain. This gives the possibility to replace heuristically parameterized operators with learned data-optimal operators.
With the last experiment we show the potential of the framework to elevate research, as we present a novel way to design a network with just a few lines of code that mimics a iterative reconstruction and, intrinsically, is a starting point to learn regularizers based on the capability of deep neural networks \cite{hammernik2018learning}. 

We choose to implement the projector and back-projector as a unmatched projector pair. The implications of unmatched pairs are already analyzed in the context of iterative reconstruction by Zeng et al.~\cite{unmatched}. Zeng et al. concluded that unmatched-pairs can be beneficial due to the algorithmic speedup, while the convergence of the algorithm has to be kept in mind. While we have not noticed negative effects on the training process in our experiments\cite{sybenctmeeting,sybengcpr}, we want to investigate the implications of unmatched-projector pairs to the training procedure in future work. 

As the combination of deep neural networks and CT reconstruction can, especially in the 3D case, easily exceeds the GPU's memory, the provided algorithms allow the user a trade-off choice between computational- and memory efficient implementations. Furthermore, the concept of the framework enables a problem-specific solution, since the algorithms and the gradients can be changed by the user at any time. Additionally, as the core of PYRO-NN is an extension of the existing Tensorflow build process, every known operator which allows the calculation of sub-gradients can easily be modeled as a Tensorflow layer. Besides the actual CUDA implementation, there is only the need of an information flow control class following the Tensorflow API guidelines. Therefore, the setup allows an easy extension towards other frameworks like PyTorch as only the information flow control class has to be adjusted, while the core implementation of the known operator stays untouched.

We provide the known operators for CT reconstruction on CUDA level with the respective necessary tools like filters and weights on Python level. Nevertheless, the framework design allows an easy extension to other fields, e.g., magnetic resonance imaging (MRI) and many more. With the increasing amount of publications being supplemented by open-source reference implementations,
we believe that with help of the community  PYRO-NN can grow beyond the application on CT reconstruction.

\section{Conclusion and Outlook}
PYRO-NN is an open-source software framework developed to elevate the use of known operators within neural networks to transfer the power of deep learning to medical image reconstruction. The framework provides state-of-the-art CT reconstruction algorithms within the Tensorflow deep learning environment, supported by the necessary tools for the reconstruction pipeline. 

In this paper, we show the concept of embedding known operators by means of CT reconstruction and underline with experiments comparable to existing algorithms and the applicability of these algorithms in the deep learning framework Tensorflow. The generic design of the framework makes it very easy to extend it to other modalities. We hope that our open-source framework will encourage other groups to join these efforts making the framework a valuable element in the deep learning medical image reconstruction field. 

The main objective of the framework is to enable the community to use CT reconstruction algorithms in end-to-end neural networks and to elevate the research in medical image reconstruction.

The software package is available under \url{https://github.com/csyben/PYRO-NN} and \url{https://github.com/csyben/PYRO-NN-Layers}.
\section*{Acknowledgment}
The research leading to these results has received funding from the European Research  Council  (ERC)  under  the  European  Union’s  Horizon  2020  research and innovation program (ERC Grant No.  810316). Additional financial support for this project was granted by the Emerging Fields Initiative (EFI) of the Friedrich-Alexander University Erlangen-N\"urnberg(FAU). 
\section*{References}
\addcontentsline{toc}{section}{\numberline{}References}
\vspace*{-20mm}





\bibliography{./deep_ct_reconstruction_paper.bbl}      



\bibliographystyle{./medphy.bst}    


\end{document}